[Title Page]

# A Three-phase Augmented Classifiers Chain Approach Based on Co-occurrence Analysis for Multi-Label Classification

Pengfei Gao, Dedi Lai, Lijiao Zhao, Yue Liang, Yinglong Ma[*]

School of Control and Computer Engineering, North China Electric Power University, Beijing 102206, China.

Email: yinglongma@gmail.com; yinglongma@ncepu.edu.cn

* Correspondence author

Correspondence information:
Full name: Yinglong Ma
Affiliation: School of Control and Computer Engineering, North China Electric Power University, Beijing 102206, China
Email address: yinglongma@ncepu.edu.cn
Telephone number: +86 10 61772643

# A Three-phase Augmented Classifier Chains Approach Based on Co-occurrence Analysis for Multi-Label Classification

Pengfei Gao, Dedi Lai[*], Lijiao Zhao, Yue Liang, Yinglong Ma[†]

School of Control and Computer Engineering, North China Electric Power University,

Beijing 102206, China.

**Abstract**

As a very popular multi-label classification method, Classifiers Chain has recently been widely applied to many multi-label classification tasks. However, existing Classifier Chains methods are difficult to model and exploit the underlying dependency in the label space, and often suffer from the problems of poorly ordered chain and error propagation. In this paper, we present a three-phase augmented Classifier Chains approach based on co-occurrence analysis for multi-label classification. First, we propose a co-occurrence matrix method to model the underlying correlations between a label and its precedents and further determine the head labels of a chain. Second, we propose two augmented strategies of optimizing the order of labels of a chain to approximate the underlying label correlations in label space, including Greedy Order Classifier Chain and Trigram Order Classifier Chain. Extensive experiments were made over six benchmark datasets, and the experimental results show that the proposed augmented CC approaches can significantly improve the performance of multi-label classification in comparison with CC and its popular variants of Classifier Chains, in particular maintaining lower computational costs while achieving superior performance.

[*] The authors Dedi Lai and Pengfei Gao contributed equally to this work (co-first author)
[†] Corresponding author. Tel.: +86 61772643. E-mail address: yinglongma@ncepu.edu.cn

*Keywords:* Multi-Label Classification; Classifier Chains; Co-occurrence Analysis; Trigram; Label Correlation.

**1. Introduction**

In the last decades, multi-label classification (MLC) has received increasing attention in the machine learning community [1], and has been widely applied in the fields of text classification [2-4], media content tagging [5], online processing [6], and protein and genome prediction [7], etc. For examples, there are many multi-label text classification tasks in the real-world applications where a large number of quite close category labels are organized, such as Open Directory Project [‡], Medical Subject Headings [§], the library and patent classification scheme [**], Wikipedia topic classifications[††], and social media websites [8], and so on.

Multi-label classification is to train and learn a model that assigns a subset of labels to each instance in the sample set, different from the traditional multi-class classification that assigns just one label to each instance [9, 10]. For example, in some text classification tasks, a newspaper article concerning the news of Messi possibly can be respectively classified into the sports news and entertainments news. The underlying fact residing in MLC is that a sample often has multiple different attributes corresponding to labels/tags. In order to accurately predict all the possible labels for each sample, a typical multi-label classification is needed to handle the main two-fold problems [11]. On one hand, the number of labels assigned or predicted to each instance is variable during the course of training and prediction due to that different samples often have different attributes [12]. This makes it challenging for many of the existing multi-label classification algorithms to achieve high accuracy classification performance when encountering a large number of labels. On the other hand, there often is latent interdependency and correlation between labels. It is crucial for multi-label classifiers to model and exploit the dependencies and correlations between labels for improving performance. For example, *Brazil* and *Football* are two labels. If a news sample is assigned the *Brazil* label, then there will be a very high probability that it is assigned the *Football* label. Current research on MLC is mainly driven by the idea that optimal predictive performance cannot be achieved without modeling and exploiting dependencies and correlations between labels [13-21].

---

[‡] https://www.dmoz-odp.org/
[§] https://meshb.nlm.nih.gov/treeView
[**] https://www.loc.gov/aba/cataloging/classification/
[††] https://en.wikipedia.org/wiki/Portal:Contents/Categories

Classifier chains (CC) [22] is a well-known MLC method with great popularity due to its simple idea and good performance [14], even though it has been introduced only lately. The CC method selects an order on the label set as a chain of labels, and then a binary classifier for each label is trained by incorporating the predicted results on the classifiers of the preceding labels in the label order as additional inputs. An ensemble of classifier chains (ECC) makes it perform particularly well as it models label correlations at acceptable complexity. However, the CC based methods often adopt the randomly generated order of labels [23, 24]. They are generally difficult to approximate the underlying dependency in the label space, and often suffer from the problems of poorly ordered chain and error propagation. Therefore, it has become a challenging problem for improving MLC performance to model the underlying correlations between labels and further take full advantages of these correlations at acceptable complexity.

Towards the goal of modeling and exploiting the underlying correlations between labels at acceptable complexity for MLC, in this paper, we present a co-occurrence analysis based augmented Classifiers Chain approach for multi-label classification by optimizing the order of labels. The contributions of this paper are as follows.

First, we present a three-phase augmented Classifier Chains framework for multi-label classification.

Second, we propose a co-occurrence matrix method to model the underlying correlations between a label and its precedents in a chain of labels.

Third, we propose two different strategies (GOCC and TOCC) to approximate the underlying dependency in the label space for MLC.

Extensive experiments were made, and the experimental results show that our augmented CC methods can significantly improve the performance of MLC, in particular maintaining lower computational costs while achieving superior performance.

This paper is structured as follows. Section 1 is the introduction. In Section 2, we will discuss the related work about variants of CC. Section 3 is the preliminaries about MLC and CC. In Section 4, we give the overview of framework and define the co-occurrence matrix. Sections 5 and 6 respectively propose the methods GCC and NCC. Section 7 is the experiments and analysis. Section 8 is the conclusion.

## 2. Related Work

In the last years, multi-label classification (MLC) has been deeply explored, and many MLC approaches have been proposed. Generally speaking, existing MLC methods can be roughly grouped into two major categories [25,26]: Algorithm Adaptation methods and Problem Transfer methods. The Algorithm Adaptation methods handle MLC learning problem by enhancing the existing machine learning models, such as MLC based on k-Nearest Neighbor (kNN) [16,26], label sorting [17], and information theory [17,18,27,28]. With the increasing attention of neural networks and deep learning, many MLC methods based on neural network models have been proposed, such as multi-layer perceptron (MLP) [7], radial basis functions, extreme learning machine, and deep neural networks [20, 29,30], etc. A BPMLL method was proposed to use neural networks for MLC, where a pairwise ranking error and hyperbolic tan activation is used to train the neural network. A cross-entropy based loss function and sigmoid activation with CNN was proposed for MLC based on end-to-end learning, improving the convergence speed and overall performance. A method OSML-ELM was proposed converts the label set from bipolar to unipolar representation in order to solve multi-label classification problems [6]. All these existing Algorithm Adaptation methods have illustrated promising potentials for MLC. However, the MLC methods based on Algorithm Adaptation often need to build more complex learning models for model training and feature representation of instances and labels, and have relevantly high complexity. Especially for the complex neural networks based CC methods, the complexity for model training will tremendously increase with the growth of instances and labels, so it remains unknown for applying them to situations where MLC training models need to be obtained as soon as possible, such as the edge computing.

The Problem Transfer methods mainly deal with transforming MLC problems into multiple sub-problems to be integrated. For example, an MLC task can be reduced to the binary classification problem for each label. The typical method is Binary relevance [31], where an MLC classification task is decomposed into *N* binary classification tasks, where *N* is the number of labels. BR has illustrated a good performance, but fails to exploit inter-class correlations that possibly are very helpful in boost performance. Classifier Chains is another typical Problem Transfer method introduced lately. It selects a label order as a chain of labels, and then trains a binary classifier for each label by incorporating the predicted results on the preceding classifiers in the label order as additional inputs. An obvious advantage of CC is that it takes into account the label correlations while maintaining acceptable computational complexity. However, CC is generally difficult to approximate the underlying dependency in the label space, and often suffer from the problems of poorly ordered chain and error propagation [45-5].

Of course, there are some approaches that exploit the CC based MLC by leveraging the interdependency and correlations [5, 22, 32, 33]. A polytree augmented classifier chains method (PCC) [5] was proposed to approximate the underlying dependency by modeling reasonable conditional dependence between labels over attributes, but it has theoretical complexity (NP-hard), whose computation cost on prediction increases exponentially in number of labels. There are also some extensions of CC such as probabilistic classifier chains (PCC) [32] and Bayesian Chain Classifiers (BCC) [33]. They still need a higher training cost due to the complexity residing in probabilistic models and Bayesian networks. In contrast, Ensembles of Classifier Chains (ECC) is a well-known method that uses a different sample of the training data to train each member of the ensemble, and increases predictive performance over CC by effectively using a simple voting scheme to aggregate predicted relevance sets of the individual CCs. An obvious advantage of ECC is that it performs particularly well as it can model label correlations at acceptable complexity [23, 24], but it is difficult to approximate the underlying dependency in the label space, and often suffers from the problems of poorly ordered chain and error propagation due to that its input is still the randomly generated order of labels [13].

Different from the work above, we present a co-occurrence analysis based augmented Classifier Chains approach for multi-label classification by optimizing the order of labels. A co-occurrence matrix method is used to model the underlying correlations between labels, and two different strategies of augmented Classifier Chains (GOCC and NOCC) are presented to approximate the underlying dependency in the label space for MLC.

## 3. Preliminaries

### 3.1 Multi-Label Classification

Let $X \subseteq \mathcal{R}^k$ be the $k$-dimensional instance input feature space, and $Y=\{l_1, l_2, \ldots, l_q\}$ be the set of labels. A sample set $D$ of training data consists of $n$ instances, denoted as $D = \{(x_i, y_i)\}_{i=1}^n$, where for each instance $(x_i, y_i) \in D$, $x_i=(x_{i,1}, x_{i,2},\ldots, x_{i,k}) \in X$ is a $k$-dimensional feature vector, and $x_{i,j}$ denotes the $j$-th element of feature vector $x_i$. We use $y_i=(y_{i,1}, y_{i,2},\ldots, y_{i,q}) \in \{0,1\}^q$ to represent a $q$-dimentional label vector, where $y_{i,j}=1$ and $y_{i,j}=0$ indicate that label $l_j$ is relevant and irrelevant to $x_i$, respectively. Let $Y_i \subseteq Y$ be the set of labels relevant to $x_i$, and we have $Y_i=\{l_j \mid y_{i,j}=1, 1 \leq j \leq q\}$.

The task of multi-label classification is to find an optimal classifier $f: X \rightarrow \{0,1\}^q$ that assigns $y_i$ to each instance $x_i$. In the context of BR [31], a classifier $f$ is comprised of $q$ binary classifiers $f_1, f_2, \ldots, f_q$. Each binary classifier $f_j: X \rightarrow \{0,1\}$ can be induced based on its relevancy (irelevancy) to $l_j$ from a derived binary training set $D_j = \{(x_i, y_{i,j})\}_{i=1}^n$, where $D_j$ is derived by transforming each instance $(x_i, y_i) \in D$ into a binary training instance $(x_i, y_{i,j})$ with respect to label $l_j$. For an instance $x'$ whose label is unknown, each classifier $f_j$ is queried to predict its associated label set $Y'=\{l_j \mid f_j(x')=1, 1 \leq j \leq q\}$.

### 3.2 Classifier Chains

Classifier Chains (CC) is a well-known MLC method that is on the basis of BR to overcome the limitation of BR that BR ignores the correlations between labels in the training data, and therefore achieves a higher predictive performance. The model realizes the series-type join of the classifiers by adding the results of the preceding classifiers to the current classifier.

Specifically, CC first randomly generates an order of label in the chain of classifiers, denoted as $Y=\{l_1, l_2, …, l_q\}$, and then CC trains a set of binary classifiers $f_1, f_2, …, f_q$ in the order of the chain of classifiers.

In the training phase, each binary classifier $f_j$: $X \rightarrow \{0,1\}$ can be induced based on its relevancy (irelevancy) to the previous $j$-1 labels $l_1, l_2, ..., l_{j-1}$ as well as the current label $l_j$ from a derived binary training set $D_j = \{(x_i, y_{i,1},..., y_{i,j-1}, y_{i,j})\}_{i=1}^{n}$, where each instance in the training set $D_j$ is derived by the corresponding instance $(x_i, y_i) \in D$.

In the testing phase, it predicts the value $f_j(x^*)$ of an unseen instance $x^*$ in a greedy manner. Each classifier $f_j$ ($1 \leq j \leq q$) is queried to predict the associated label set $Y^*$ of instance $x^*$, where $Y^*=\{l_j \mid f_j(x^*)=1, 1 \leq j \leq q\}$.

CC obviously utilizes the interdependency relationships between labels, but it is still very sensitive to the order of a chain of classifiers due to the fact that the original randomly generated label order of a chain cannot effectively avoid the risk of error propagation in advance. Therefore, it has become a key problem how to select the most approximate order of a chain of classifiers for ensuring high accuracy of MLC.

## 4. The Proposed Framework and Co-occurrence Matrix

### 4.1 Overview of Framework for MLC

In this paper, our approach framework of the optimal order based CC augmented MLC can be divided into the following three sequential phases: Co-occurrence Analysis, Optimal Order of Labels, and MLC Training and Testing Based on CC. The overview of our framework is shown in Figure 1.

In the first phase Co-occurrence Analysis, we propose a co-occurrence rate (CR) matrix based approach to model the interdependency and correlations between labels, through which we can determine the previous two labels in the order of a chain for CC, which will be discussed in subsections 4.2 and 4.3.

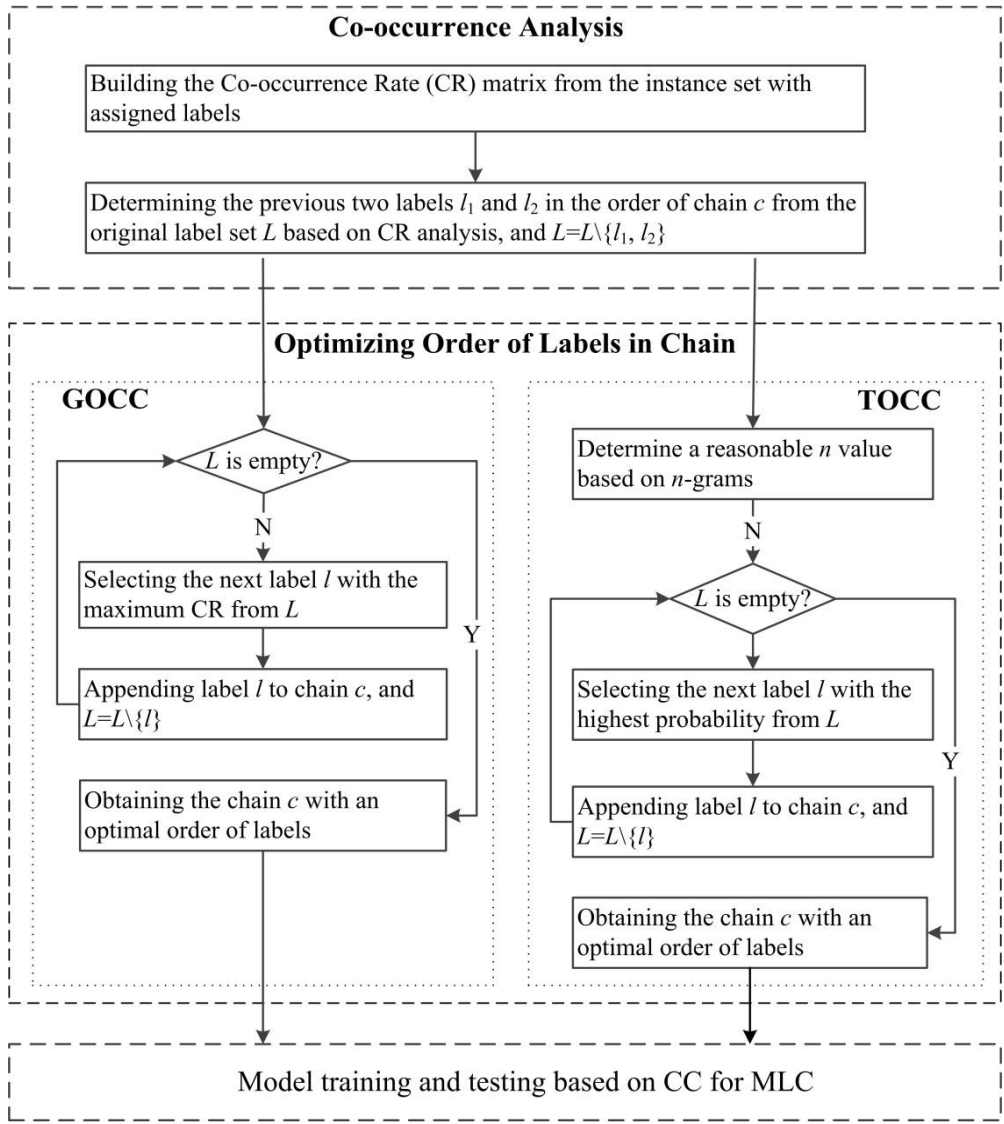

Fig. 1 The three-phase based framework for CC based MLC

The second phase Optimal Order of Labels is to generate a complete label chain in the optimal order of labels for MLC based on CC, where two strategies for generating an optimal order of labels are introduced, including Greedy Order Classifiers Chain (GOCC) and N-gram Order Classifiers Chain (NOCC). Different from most of existing strategies for improving MLC performance based on CC [22], in this paper we mainly focus on generating an optimal order of labels based on the co-occurrence rate matrix based analysis, which will be specifically discussed in sections 5 and 6, respectively.

The last phase is just to perform the model training and testing based on CC model for MLC, using the chain of labels generated in the previous phase.

4.2 Co-occurrence rate Matrix and Analysis

In this paper, we propose a co-occurrence rate (CR) matrix based approach to model the interdependency and correlations between labels. Co-occurrence analysis provides a way to quantitatively measure the correlation degree of the potential relationship between two elements by calculating the frequency of the occurrence of two elements together[34], which has been currently applied in word embedding technique [35]. The construction of a CR matrix is as follows.

Let $D=\{(x_i, y_i)\}_{i=1}^{n}$ be the training set, and $Y=\{l_1, l_2, \ldots, l_q\}$ be the set of labels. Let $y_i=(y_{i,1}, y_{i,2},\ldots, y_{i,q})$, and $Y_i \subseteq Y$ be the set of labels relevant to $x_i$, i.e. $Y_i=\{l_j \mid y_{i,j}=1, 1 \leq j \leq q\}$. Furthermore, we use $S_i=\{(x_j, y_j) \mid (x_j, y_j) \in D, y_{j,i}=1\}$ to denote the set of instances relevant to label $l_i$, and correspondingly use $\overline{S_i}=\{(x_j, y_j) \mid (x_j, y_j) \in D, y_{j,i}=0\}$ to denote the set of instances irrelevant to $l_i$. Thus, $S_i \cap S_j=\{(x_k, y_k) \mid (x_k, y_k) \in D, y_{k,i}=1 \text{ and } y_{k,j}=1\}$.

In this paper, the CR matrix $M$ for modeling label correlation is a $q \times q$ matrix, which is derived from the training set $D$, where the element $M_{i,j}$ of the $i$-th row and the $j$-th column in matrix $M$, can be defined in equation (1).

$$M_{i,j} = \frac{|S_i \cap S_j| \cup |\overline{S_i} \cap \overline{S_j}|}{n} \quad (1)$$

Where $|S|$ denotes the total number of elements in set $S$, and $n$ is the total number of labels in the training set $D$.

Due to the lower probability in which each instance is relevant to a large number of labels at the same time, the generated co-occurrence rate matrix will be very sparse. For this reason, in equation (1) we also take into account both the situations where two labels are simultaneously relevant to each instance, or simultaneously irrelevant to each instance. We believe that both the situations can better reflect the correlation between two labels. The complexity for calculating each $M_{i,j}$ of the CR matrix is $O(|D|)$ at the worst case. Due to that it is meaningless to calculate the co-occurrence of the same label, and therefore the total complexity of building the CR matrix is $O(|D| \times (|Y|-1)^2)$.

Based on the definition of co-occurrence rate matrix, we give an example of the co-occurrence rate matrix $M$ shown in table 1, where the set of labels is $Y= \{l_1, l_2, l_3, l_4, l_5\}$. In $M$, each cell $M_{i,j}$ quantitatively represents the co-occurrence based correlation between labels $l_i$ and $l_j$ in the form of proportion, so as to facilitate the subsequent label correlation exploration.

Tab. 1 An example of co-occurrence ratio matrix $M$

|       | $l_1$ | $l_2$ | $l_3$ | $l_4$ | $l_5$ |
|-------|-------|-------|-------|-------|-------|
| $l_1$ | —     | 0.255 | 0.063 | 0.045 | 0.035 |
| $l_2$ | 0.255 | —     | 0.232 | 0.073 | 0.063 |
| $l_3$ | 0.063 | 0.232 | —     | 0.246 | 0.051 |
| $l_4$ | 0.045 | 0.073 | 0.246 | —     | 0.135 |
| $l_5$ | 0.035 | 0.063 | 0.051 | 0.135 | —     |

4.3 Determination of the Previous Two Labels in Chain

We determine the previous two labels for the optimal order of the chain of classifiers by analyzing the CR matrix $M$. We have to traverse each row in $M$, and find the cells with the maximum value. We present a method to pairwise compare each label pair $(l_i, l_j)$ by $M_{i,j}$ in the CR matrix for selecting the pairwise labels with the maximum $M_{i,j}$ value as the previous two labels, which can be defined in equation (2).

$$H = \arg\max_{1 \leq i, j \leq q}\{qut(M_{i,j})\} \qquad (2)$$

Where notation $H$ denotes the set of number pairs with the maximum CR value of the form $(i, j)$ referring to a label pair $(l_i, l_j)$, and the notation $qut(M)$ represents the quasi-upper triangular matrix of $M$. We notice that the CR matrix $M$ is symmetric, thus we only need to use its quasi-upper triangular matrix for speeding up element traversing.

It is crucial to decide the head label in the order of chain. The basic criterion is to rank the labels with higher CR values as previous as possible, which will make the results of the preceding classifiers be efficiently exploited for the current classifier based classification. Here, two possible steps need to be taken into accounts.

First, assume that $H$ only contains just one label pair. Let $H=\{(i, j)\}$, we need to compare if $\max_{1 \leq s \leq q, s \neq j}\{M_{i,s}\} \geq \max_{1 \leq s \leq q, s \neq i}\{M_{j,s}\}$. If it does, then $l_i$ and $l_j$ are respectively the first and the second labels in the order of chain, and vice versa.

Second, if $H$ contains two label pairs, suppose that $H=\{(i, j), (s, t)\}$ without loss of generality,. We need to respectively calculate the CR values between their two endpoint labels and the other labels. We compute $\max_{1 \leq r \leq q, r \neq i, r \neq j}\{M_{i,r}\}$ and $\max_{1 \leq r \leq q, r \neq i, r \neq j}\{M_{j,r}\}$ for pair $(l_i, l_j)$, and $\max_{1 \leq r \leq q, r \neq s, r \neq t}\{M_{s,r}\}$ and $\max_{1 \leq r \leq q, r \neq s, r \neq t}\{M_{t,r}\}$ for pair $(l_s, l_t)$. Then, we need to judge if $\max\limits_{1\leq r\leq q, r\neq i, r\neq j}\{M_{i,r}\} + \max\limits_{1\leq r\leq q, r\neq i, r\neq j}\{M_{j,r}\} > \max\limits_{1\leq r\leq q, r\neq s, r\neq t}\{M_{s,r}\} + \max\limits_{1\leq r\leq q, r\neq s, r\neq t}\{M_{t,r}\}$ holds.

If it does, pair (*i*, *j*) will be used. Once a label pair is finally determined, we can further decide which label in the pair will respectively be the first and the second in the order of chain by step 1.

Once we determine the previous labels in the order of chain, we will use the following two strategies to select the subsequent order of labels. They are called Greedy Order of Labels for Classifiers Chain and Trigram Order of Labels for Classifiers Chain.

## 5. Greedy Order of Labels for Classifiers Chain (GOCC)

We present a greed algorithm based strategy to determine the order of labels of chain. It is called Greedy Order of Labels for Classifiers Chain (GOCC). Because the previous labels have been determined, we only continue to decide the remaining order of chain. The GOCC strategy needs to traverse all elements at each row, as shown in Algorithm 1. In algorithm 1, lines 1 to 4 are to put the pair (*s*, *t*) is put in the order as the previous two labels, and make some initializations. In lines 5 to 10, we select a subsequential label for the current label in a greedy manner by making maximal the co-occurrence rate of the subsequential label.

**Algorithm 1**. Greedy Order of Labels for Classifiers Chain (GOCC)

**Input**: (1) $M = \{M_{i,j}\}_{i=1, j=1}^{i=q, j=q}$ :the CR matrix where *q* is the total number of labels;
    (2) (*s*, *t*) ∈*H*: the determined pair of previous labels in the order of chain.
**Output**: *L*[ ]: the order of labels.
**Begin**
    //the pair (*s*, *t*) is put into the order
1. $L[0] \leftarrow s$;
2. $L[1] \leftarrow t$;
3. $T \leftarrow \{s, t\}$;
4. $c \leftarrow 1$; //an index recording the number of a label.

5.  **while** $c \leq q\text{-}1$ **do**
    // select a label with the maximum CR value at row $L[c]$
6.      $k \leftarrow \max\limits_{1 \leq j \leq q, j \notin T}\{M_{L[c],j}\}$ ;
7.      $L[c+1] \leftarrow k$;
8.      $T \leftarrow T \cup \{k\}$;
9.      $c \leftarrow c+1$;
10. **endwhile**
11. **return** $L[]$;
**End**

By performing algorithm 1 above, we will re-rank all the labels from the original set $Y$ of labels accorging to their numbers, i.e., the optimal order of labels in the chain is $l_{L[0]} \rightarrow l_{L[1]} \rightarrow \ldots \rightarrow l_{L[q-1]}$. The complexity of the algorithm GOCC including the construction complexity of CR matrix is only $O(|D| \times |Y|)$ at the worst case.

**6. Trigram Order of Labels for Classifier Chain (TOCC)**

In this section, we will propose an $n$-gram model based strategy for determining the order of labels of chain. The $n$-gram model is language model and currently has been used as an efficient method for capturing word order in probabilistic distribution between words before and after [32]. It is crucial for the $n$-gram model to determine a reasonable $n$ value that confines the size of a sliding window representing the context of words [36]. The selection of $n$ value has a direct influence to model complexity. The labels in this paper can be treated as the words in the $n$-gram language model. The empirical validation was made in this paper, which indicates that the trigram model (i.e., the $n$-gram where $n=3$) is most efficient to MLC. So our $n$-gram model based strategy is called Trigram Order of Labels for Classifiers Chain (TOCC).

A Three-phase Augmented Classifiers Chain Approach for Multi-Label ClassificationIn the *n*-gram language model, we suppose that the order of labels to be constructed currently has a total of *m* words: $w_1, w_2, \ldots, w_m$, and the current word is only related to the previous *n*-1 ($n \ll m$) words in front of it. The *n*-gram based probability distribution of the word sequence $w_1, w_2, \ldots, w_m$ is generated as shown in equation (3).

$$P(w_1,...,w_m) = P(w_1)P(w_2 \mid w_1)...P(w_m \mid w_{m-n+1}, \cdots, w_{m-1}) \qquad (3)$$

In equation (3), the different values of *n* have different influence on calculating the probability distribution of the whole sequence. If *n* is too large, then the time complexity will be sharply increased, which makes the original task of capturing word order more difficult. Just because of this, many applications based on the *n*-gram model beforehand need to empirically select a most suitable *n* value. In the subsection 7.4.1 of this paper, an empirical validation was made where different *n* values are selected to verify which *n* value is most efficient. The experiment results show that there will be a best performance when *n*=3. So in this paper, our strategy (TOCC) is implemented based on the trigram language model.

For generating an optimal order of labels for Classifiers Chain, let *s* represent an order of labels of a chain with the length *q*. We define the trigram based probability distribution in an order of labels of the chain, which is shown in equation (4). By the trigram model, we can convert the probability of chain *s* into the product of conditional probability.

$$P(s) = P(l_1, l_2, ..., l_q) = P(l_1)P(l_2 \mid l_1)P(l_3 \mid l_1, l_2) \cdots P(l_q \mid l_{q-2}, l_{q-1}) \qquad (4)$$

As mentioned in the previous section, we have determined the first two labels for the optimal order of labels. What we should do in the following is to determine the remaining order of other subsequent labels. We suppose that $s_{i-1}=l_1, l_2, \ldots, l_{i-2}, l_{i-1}$ is an optimal suborder of labels from the determined head label to label $l_{i-1}$, where $i \geq 3$. The conditional probability $P(l_i| l_{i-2}, l_{i-1})$ of $l_i$ based on the trigram model can be defined in equation (5).

$$P(l_i | l_{i-2}, l_{i-1}) = \frac{|S_i \cap S_{i-1} \cap S_{i-2}|}{|S_{i-1} \cap S_{i-2}| + 1} \qquad (5)$$

Where the sets $S_{i-2}$, $S_{i-1}$ and $S_i$ are respectively to represent the instance sets relevant to labels $l_{i-2}$, $l_{i-1}$ and $l_i$, and each $S_i = \{(x_j, y_j) \mid (x_j, y_j) \in D, y_{j,i}=1\}$.

It is not difficult to find that the condition probability $P(l_i| l_{i-1}, l_{i-2})$ of the subsequent label $l_i$ of must be maximized in order to maximize the value of $P(s_i)$. Let $S$ be the label set contained in the ordered sequence $s_{i-1}=l_1, l_2, \ldots, l_{i-1}$, and let $Y'=Y \setminus S$, then we select a label $l_i$ as a subsequent label from $Y'$ according to equation (6).

$$l_i = \arg\max_{l_i \in Y} \{P(l_i | l_{i-2}, l_{i-1})\} \qquad (6)$$

**Algorithm 2.** Trigram Order of Labels for Classifiers Chain (TOCC)

**Input**: (1) $D = \{(x_i, y_i)\}_{i=1}^{n}$: the instance set with the label set $Y=\{l_1, l_2, \ldots, l_q\}$;
   (2) $(s, t) \in H$: the determined pair of previous labels in the order of chain.
**Output**: $L[\ ]$: the order of labels.
**Begin**
   //the pair $(s, t)$ is put into the order
1. $L[0] \leftarrow s$;
2. $L[1] \leftarrow t$;
3. $Y \leftarrow Y \setminus \{l_{L[0]}, l_{L[1]}\}$;
4. $c \leftarrow 2$; //an index recording the number of a label.
5. **while** $Y \neq \emptyset$ **do**
   // select a label with the maximum conditional probability
6.    $B \leftarrow \{(x_k, y_k) \in D \mid y_{k,L[c-1]}=1, y_{k,L[c-2]}=1\}$;
7.    $max \leftarrow 0$;
8.    **foreach** $l_j \in Y$ **do**
9.       $U \leftarrow \{(x_k, y_k) \in P \mid y_{k,j}=1\}$
10.      $p \leftarrow |U|/(|B|+1)$;
11.      **if** $p > max$ **then**
12.         $max \leftarrow p$;
13.         $L[c] \leftarrow j$;
14.      **endif**

```
15.        endfor
16.     Y←Y\{ l_{L[c]} };
17.     c ← c+1;
18. endwhile
19. return L[ ];
End
```

Algorithm 2 is the pseudocode for the strategy TOCC. We only need to determine the remaining order of chain due to that the previous labels have been determined. In algorithm 2, lines 1 to 4 are to put the pair ($s$, $t$) is put in the order as the previous two labels, and make some initializations. In lines 5 to 18, we select a subsequential label for the current label by making maximal the conditional probability of the subsequential label. In lines 8 to 15, the label with the highest probability $P(l_c | l_{c-2}, l_{c-1})$ is selected as the subsequent label of the previous determined sub-order. The computational complexity of algorithm 2 is $O(|D|\times|Y|^2)$ at the worst case. In the real world, the number $|Y|$ of labelsis ften far less than the number $|D|$ of instances. Algorithm 2 almost have a linear compexity with respect to the number of the trained instances.

Based on the order of labels of chain determined by the algorithms GOCC and TOCC, Classifiers Chain model can be further performed and used for MLC.

**7. Experiment Evaluation and Analysis**

The main objective of the experiments is to evaluate the efficiency of algorithms GOCC and TOCC proposed in the previous sections.

7.1 Datasets Information

In our experiments, we used the six benchmark datasets including Yeast, Enron, Emotion, Slashdot-F, CAL500 and Scence, which were obtained from the literature [3] and some websites MuLan[‡‡] and MEKA[§§]. These datasets include text, pictures, and other types of data. The information details of the dataset are shown in table 2, where LCard refers to the average number of labels per instance.

Tab. 2 Description of datasets

| Name | Instances | Features | Labels | LCard |
|---|---|---|---|---|
| Emotions | 593 | 72 | 6 | 1.879 |
| Enron | 1702 | 1001 | 53 | 3.378 |
| Yeast | 2417 | 103 | 14 | 4.237 |
| Slashdot-F | 1460 | 1079 | 22 | 1.18 |
| CAL500 | 502 | 68 | 174 | 26.044 |
| Scence | 2407 | 294 | 6 | 1.074 |

All the algorithms in the experiments were implemented by python and its sklearn package. In the selection of base classifier, in this paper we adopted SVM, whose kernel function is the Gaussian function and the penalty parameter was set C=100. All algorithms use the same parameters in the base classifier to avoid the difference in the base classifier which will possibly affect the effect of order optimization. In these algorithms, ECC is trained with four random sequences. ML-KNN sets K=20 as the number of the nearest samples. In TSVA, the threshold was set to 0.01 on the datasets Slashdot-F and Enron due to that both them include more features, and 0.15 on the remaining datasets.

7.2 Baselines

Many CC based MLC approaches have recently been proposed. The existing approaches besides the original Classifiers Chain [22] typically include Binary

---
[‡‡] Available at http://mulan.sourceforge.net/datasets-mlc.html
[§§] Available at http://waikato.github.io/meka/datasets/

A Three-phase Augmented Classifiers Chain Approach for Multi-Label Classification

Relevance (BR) [31], Ensembles of Classifiers Chains (ECC) [22], the multi-label K-nearest neighbor algorithm (ML-KNN) [26], Two stage voting architecture algorithm(TSVA) [37] and the Random k-labelsets algorithm (RAKEL) [23]. In this paper, these mentioned algorithms will be used as the baselines for MLC performance comparison and analysis. The information of these algorihtms is shown in Table 3.

Tab. 3 Description of baseline algorithm

| Name of MLC Related to CC | Abbreviation | Reference |
|---|---|---|
| Binary Relevance | BR | [31] |
| Classifier Chains | CC | [22] |
| Ensemble of Classifier Chains | ECC | [22] |
| Multi-Label K- Nearest Neighbor | ML-KNN | [25] |
| Two Stage Voting Architecture | TSVA | [37] |
| RAndom K-labELsets | RAKEL | [23] |

7.3 Performance Indexes

In this paper, we used many typical performance indexes for experimental evaluation because a single performance index is difficult to reflect the complete performance. Therefore, we adopt the accuracy, F1-score and Hamming loss [38] of multi-label classification as performance evaluation indexes.

(1) Accuracy

$$Accuracy = \frac{1}{|D|} \sum_{i=1}^{|D|} \frac{|Y_i \cap P_i|}{|Y_i \cup P_i|} \quad (7)$$

Where |D| represents the number of instances in dataset D. $Y_i$ and $P_i$ respectively denote the true and predicted label set of the i-th instance.

(2) F1-score

$$F1 = \frac{1}{|L|} \sum_{i=1}^{|L|} \frac{2 \times p_i \times r_i}{p_i + r_i} \quad (8)$$

Where $p_i = \frac{|Y_i \cap P_i|}{|Y_i|}$, $r_i = \frac{|Y_i \cap P_i|}{|P_i|}$. F1-score is an indicator used to measure the comprehensive performance of the classification, taking into account the correct labels and the correct and wrong examples. |L| is the total number of labels.

(3) Hamming Loss (HLoss)

$$HLoss = \frac{1}{|D|}\sum_{i=1}^{|D|}\frac{Y_i \oplus P_i}{L} \tag{9}$$

Hamming loss is perhaps the most intuitive and understandable loss function, which directly counts the number of misclassified labels, i.e., the labels irrelevant (relevant) to an instance are predicted as relevant (irrelevant). The notation $\oplus$ is an xor operation based on sets. $HLoss=0$ means that all the labels of each instance are predicted correctly.

(4) Average Accuracy

$$\overline{Accuracy} = \frac{1}{|DS|}\sum_{i=0}^{|DS|}Accuracy_i \tag{10}$$

(5) Average F1-score

$$\overline{F1} = \frac{1}{|DS|}\sum_{i=0}^{|DS|}F1_i \tag{11}$$

(6) Average Hamming Loss

$$\overline{HLoss} = \frac{1}{|DS|}\sum_{i=0}^{|DS|}HLoss_i \tag{12}$$

Where |DS| denotes the total number of datasets, $Accuracy_i$, $F1_i$ and $HLoss_i$ respectively denote the accuracy, F1-score and Hamming loss on each dataset.

(7) Average Time

$$\overline{Time} = \frac{1}{|DS|}\sum_{i=0}^{|DS|}Time_i \tag{13}$$

Where $Time_i$ represents the performing time of algorithm running on the *i-th* dataset.

We also introduce four indexes such as the normalized average time, the normalized accuracy, the normalized F1-score and the normalized average Hamming Loss in order to observe their performance differences more intuitively.

(8) Normalized Average Time

$$Time^*_j = \frac{\overline{Time}_j - \overline{Time}_{min}}{\overline{Time}_{max} - \overline{Time}_{min}} \tag{14}$$

(9) Normalized Average Accuracy

$$Accuracy^*_j = \frac{\overline{Accuracy}_j - \overline{Accuracy}_{min}}{\overline{Accuracy}_{max} - \overline{Accuracy}_{min}} \tag{15}$$

(10) Normalized Average F1-score

$$F1^*_j = \frac{\overline{F1}_j - \overline{F1}_{min}}{\overline{F1}_{max} - \overline{F1}_{min}} \tag{16}$$

Where $\overline{Time_j}$ is the average running time of the *j-th* algorithm, $\overline{Time_{min}}$ and $\overline{Time_{max}}$ are respectively the maximum and minimum average running time among these algorithms.

7.4 Experiments Comparison and Analysis

7.4.1 Determination of the n value for TOCC

Due to that the performance of *n*-gram model is closely associated with *n*, we performed our TOCC approach over these datasets by assigning *n* different integers for validating the efficacy. The accuracy and F1-score in the experiments are respectively shown in figures 2 and 3.

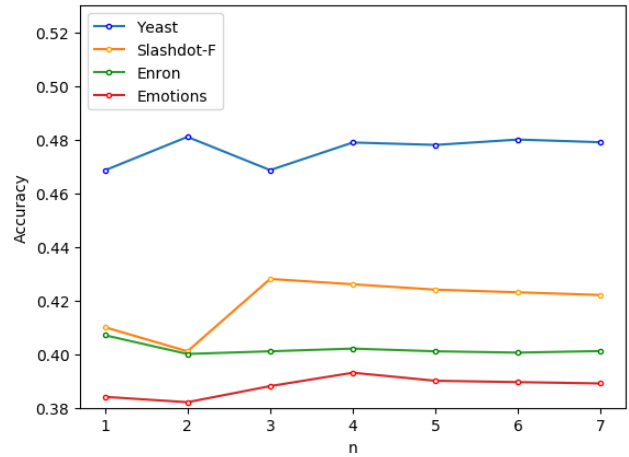

Fig. 2 Accuracy comparison over different datasets with respect to n value

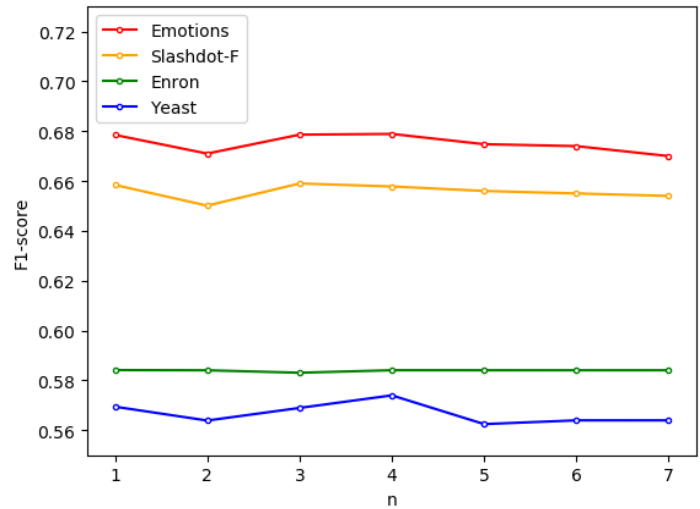

Fig. 3 F1-score comparison over different datasets with respect to *n* value

As we can see, TOCC will perform very well on the datasets when *n*=3. When *n* is assigned 1 or 2, the accuracy and F1 score for TOCC is unstable. When *n*≥3, they cannot bring more performance improvement in most cases, and the classification efficiency is basically stable. That *n* is assigned 3 in this paper is reasonable, which also illustrates the efficacy of our trigram based TOCC strategy.

7.4.2 Experiments and Analysis about Accuracy, F1 Score and Hamming Loss

We adopted the method of 5-fold cross validation for our experiments about Accuracy, F1 Score and Hamming Loss. The experimental results are shown in tables 4 to 6, where we respectively used "**0.xxxx\***", "0.xxxx*" **and** "0.xxxx" to highlight the first, second and third algorithms that have the most efficient performance among these algorithms at each row of table.

As shown in table 4, the three algorithms ECC, we find that TOCC and GOCC have illustrated the superior accuracy performance. ECC has the highest accuracy and average accuracy over almost datasets. TOCC and GOCC also achieve very good accuracy although they are very slightly lower than ECC. The accuracy performance of TOCC and GOCC is almost the same whatever we consider accuracy or average accuracy. The MLKNN method needs to further improve its accuracy performance in comparison with the other algorithms.

Tab. 4 Accuracy comparison of different algorithms

| Dataset | GOCC | TOCC | CC | BR | ECC | MLKNN | Rakel | TSVA |
|---|---|---|---|---|---|---|---|---|
| **yeast** | **0.4832*** | 0.4765 | 0.4585 | 0.4636 | 0.4751 | 0.4663 | 0.4790* | 0.4563 |
| **emotions** | 0.3789 | 0.3858* | 0.3851 | 0.3728 | **0.4018*** | 0.3338 | 0.3678 | 0.3685 |
| **enron** | 0.4026 | 0.4068 | 0.4034 | 0.4034 | **0.4280*** | 0.3513 | 0.4077* | 0.4034 |
| **Slashdot-F** | 0.4111 | **0.4239*** | 0.3945 | 0.4065 | 0.4170* | 0.2011 | 0.4049 | 0.4066 |
| **CAL500** | 0.2357* | 0.2293 | 0.221 | 0.2191 | **0.2557*** | 0.2093 | 0.2201 | 0.2194 |
| **Scene** | 0.6234* | 0.6132 | 0.5791 | 0.596 | **0.6567*** | 0.4663 | 0.6078 | 0.5936 |
| **avg_Acc** | 0.4224 | 0.4226* | 0.4069 | 0.4102 | **0.4390*** | 0.3380 | 0.4145 | 0.4079 |

From table 5, regarding the performance measurement of F1-score, we find that ECC, TOCC and GOCC also have illustrated the superior performance. TOCC has the highest average F1 score performance among these algorithms and it performs very well over almost datasets. GOCC has the same average F1 value as ECC, and it performs best over the datasets *emotions* and *CAL500*. The MLKNN method has not performed very well in comparison with the other algorithms, and still needs to be further improved for its accuracy and F1 performance.

Tab. 5 F1-score comparison of different algorithms

| Dataset | GOCC | TOCC | CC | BR | ECC | MLKNN | Rakel | TSVA |
|---|---|---|---|---|---|---|---|---|
| **yeast** | 0.5594 | 0.5717* | 0.5585 | 0.537 | 0.5574 | **0.5830*** | 0.5526 | 0.5353 |
| **emotions** | **0.6787*** | 0.6736 | 0.6563 | 0.6516 | 0.6741* | 0.6166 | 0.6649 | 0.6666 |
| **enron** | 0.5845 | **0.5866** | 0.5834 | 0.5843 | 0.5861* | 0.5269 | 0.5838 | 0.5843 |
| **Slashdot-F** | 0.6549 | 0.6563* | 0.6503 | 0.6422 | **0.6564*** | 0.5593 | 0.642 | 0.6423 |
| **CAL500** | **0.5159*** | 0.5110 | 0.5098 | 0.5088 | 0.5120* | 0.4838 | 0.5075 | 0.4985 |
| **Scene** | 0.8515 | 0.8483 | 0.8547 | 0.8547 | **0.8595*** | 0.8451 | 0.8560* | 0.8529 |
| **avg_F1** | 0.6409* | **0.6413*** | 0.6355 | 0.6298 | 0.6409* | 0.6024 | 0.6345 | 0.6299 |

From table 6, we find that the methods TSVA, BR and Rakel do the best in reducing Hamming loss. It can be seen that TSVA algorithm has the lowest Hamming loss a prominent performance in Hamming loss. The GOCC, TOCC and ECC methods have a medium level of Hamming loss. Especially, TOCC does well in reducing Hamming loss over datasets *emotions* and *enron*. MLKNN has the largest Hamming loss.

Tab. 6 Hamming loss comparison of different algorithms

| Dataset | GOCC | TOCC | CC | BR | ECC | MLKNN | Rakel | TSVA |
|---|---|---|---|---|---|---|---|---|
| yeast | 0.204 | 0.2095 | 0.2085 | **0.1988*** | 0.2115 | 0.2002 | 0.1992 | 0.1989* |
| emotions | 0.2723 | 0.2698 | 0.2743 | 0.274 | 0.2752 | **0.2656*** | 0.2783 | 0.2685* |
| enron | 0.0534 | 0.0519 | 0.0526 | 0.0519 | 0.0514* | 0.0537 | **0.0511*** | 0.0520 |
| Slashdot-F | 0.0377 | 0.0372 | 0.0439 | **0.0363*** | 0.0394 | 0.0554 | 0.0366* | 0.0364* |
| CAL500 | 0.1851 | 0.1846 | 0.1803 | 0.1680 | 0.1788 | **0.1426*** | 0.1703 | 0.1513* |
| Scene | 0.0889 | 0.0903 | 0.0867 | 0.0830* | 0.0855 | 0.2002 | 0.0858 | **0.0827*** |
| avg_hloss | 0.1402 | 0.1404 | 0.1411 | 0.1350* | 0.1403 | 0.1529 | 0.1369 | **0.1316*** |

Based on the above analysis, we can find that ECC, GOCC and TOCC perform very well over these datasets against these existing algorithms from the perspective of (average) accuracy and (average) F1 measure. TOCC has especially illustrated superior performance than GOCC. But they have a medium level of Hamming loss in comparison with TSVA, BR and Rakel. MLKNN does not perform very well over these datasets against all these indicators including accuracy, F1 and Hamming loss.

In brief, TOCC and GOCC have achieved obviously performance improvement. The underlying reason probably is that both algorithms focus on ranking the labels with the maximum CR in the front of the chain, which influences the effect of MLC. GOCC considers CR between the previous label and the current label, while TOCC considers to take the classification results of the previous two labels ($n$-1 and $n$=3) for classification on the current label. Due to that more correlations between labels are considered, TOCC performs better than GOCC.

7.4.3 Comprehensive Experiments and Analysis

In the following, we first validate the running time and the average time ($\overline{Time}$) of these algorithms over the six datasets, which is shown in figure 4 as follows.

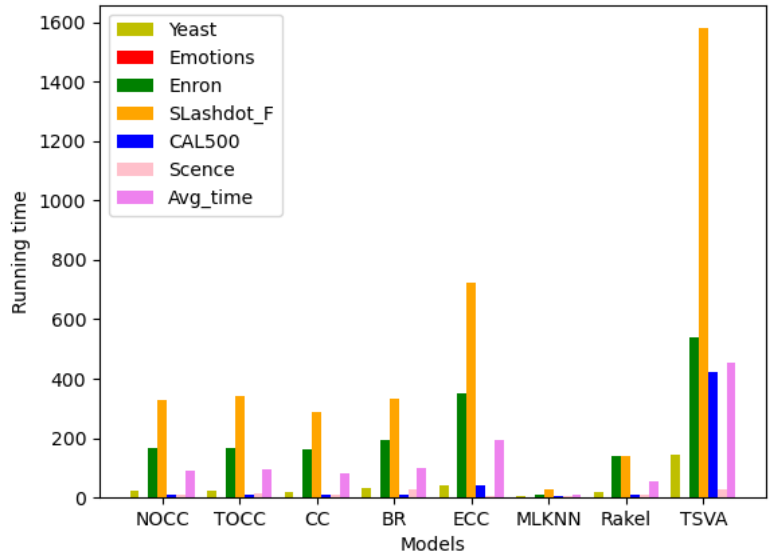

Fig. 4 Comparison of performing time and average time of different algorithms over datasets

From figure 4, we find that ECC and TSVA have the highest time cost for performing MLC over the datasets. ML-KNN and Rakel have the lowest time cost among these methods. Our TOCC and GOCC have the medium level of running time cost. However, we argue that it is difficult for completely observing the comprehensive performance to simply consider the single time dimension.

In order to completely validate and analyze the comprehensive performance of these MLC algorithms, we used some normalized indexes such as Normalized Average Accuracy ($Accuracy^*_j$), Normalized Average F1-score ($F1^*_j$) and Normalized Average Time ($Time^*_j$), which are defined in equations (14), (15) and (16). The related experiments were made and the experimental results are shown in figure 5. By observing the results in figure 5, we can further analyze the comprehensive efficiency of various algorithms more intuitively.

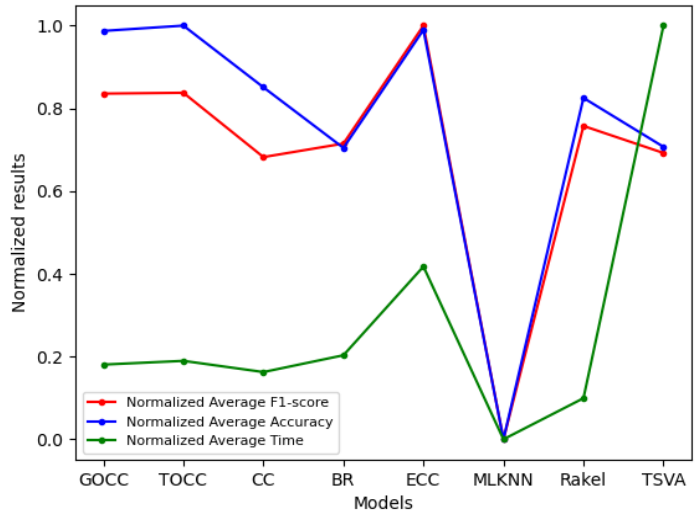

Fig. 5 Comparison of comprehensive effectiveness of different algorithms

From experimental results of the normalized indexes in figure 5, it can be observed that, first of all, TOCC and GOCC have achieved the best normalized average accuracy in comparison with these algorithms with the very low running time cost. They also have the highest F1 score in comparison with these algorithms with exception to ECC, but ECC obviously has higher normalized average time than TOCC and GOCC (the normalized average time of ECC is twice higher than TOCC and GOCC). TSVG has the highest normalized average time, which possibly makes it difficult to handle large volume of training set when we need to obtain a trained model as soon as possible for MLC. What is surprised is that ML-KNN has the lowest performance against all the three normalized indexes. If we double check tables 4, 5 and 6, we can find that ML-KNN almost illustrates the lowest performance.

In brief, from the above experiments and analysis, it is not difficult to find that our GOCC and TOCC strategies have achieved superior comprehensive performance, and can significantly improve MLC performance including accuracy, F1 score and running time against the existing popular variants of Classifier Chains, in particular maintaining computational costs while achieving superior performance. In considering the characteristics of our methods proposed in this paper, i.e., superior MLC performance to the existing MLC methods and lower complexity cost (almost linear complexity w.r.t number of instances), our approaches are rather suitable to those applications where an MLC training model needs to be acquired as soon as possible, such as object detection and recognition on the edge computing environments.

## 8. Conclusion

In this paper, we presented a co-occurrence analysis based augmented Classifier Chains approach for multi-label classification by optimizing the order of labels. We proposed a co-occurrence matrix method to model the underlying correlations between a label and its precedents. Then, GOCC and TOCC were proposed to optimize the order of labels to approximate the underlying dependency in the label space. Extensive experiments based on six benchmark datasets were made in comparison with many performance indexes, and the experimental results show that TOCC and GOCC have achieved the superior comprehensive MLC performance in comparison with these algorithms with the very low running time cost, including accuracy, F1 score and running time. The future works is to use our approaches for domain specific applications such as power text mining and fault type recognition, etc.

**Declarations**

The authors declare that they have no known competing financial interests or personal relationships that could have appeared to influence the work reported in this article. They have no conflicts of interest to declare that are relevant to the content of this article.

**References**

[1] Read J , Pfahringer B , Holmes G , et al. Classifier Chains: A Review and Perspectives. Preprint arXiv:1912.13405v2, 2020.

[2] Guo Y , Chung F , Li G . An ensemble embedded feature selection method for multi-label clinical text classification. Proceedings of IEEE International Conference on Bioinformatics & Biomedicine, pp.823-826, 2016.

[3] Yadollahi A , Shahraki A G , Zaane O R . Current State of Text Sentiment Analysis from Opinion to Emotion Mining. ACM Computing Surveys, 50(2):25, 2017.

[4] Nam J , Kim J , Mencía, Eneldo Loza, et al. Large-scale Multi-label Text Classification - Revisiting Neural Networks. Proceedings of Joint European Conference on Machine Learning and Knowledge Discovery in Databases(ECML PKDD 2014), Springer, Berlin, Heidelberg, pp.437-452, 2014.2014.

[5] Sun L , Kudo M. Polytree-Augmented Classifier Chains for Multi-Label Classification. in Proceedings of Twenty-Fourth International Joint Conference on Artificial Intelligence (AAAI2015), pp.3834-3840, 2015.

[6] Venkatesan R , Er M J , Dave M , et al. A novel online multi-label classifier for high-speed streaming data applications. Evolving Systems, 8:303-315, 2017.

[7] Zhang M L , Zhou Z H . Multilabel Neural Networks with Applications to Functional Genomics and Text Categorization. IEEE Transactions on Knowledge & Data Engineering, 2006, 18(10):1338-1351.

[8] Scholz M , Schnurbus J , Haupt H , et al. Dynamic Effects of User- and Marketer-Generated Content on Consumer Purchase Behavior: Modeling the Hierarchical Structure of Social Media Websites. Decision Support Systems, 113:43-55, 2018.

[9] Tsoumakas G , Katakis I . Multi-label classification: an overview. International Journal of Data Warehousing and Mining, 3(3):1-13, 2007.

[10] Lodhi H , Saunders C , Shawe-Taylor J. Text classification using ctring kernels. Journal of Machine Learning Research, 2(3):419-444, 2002.

[11] Tai F , Lin H T . Multilabel Classification with Principal Label Space Transformation. Neural Computation, 24(9):2508-2542, 2012.

[12] Zhang Y , Zhou Z H . Multilabel dimensionality reduction via dependence maximization. Proceedings of the 23rd national conference on Artificial intelligence (AAAI'08) pp.1503–1505, 2008.

[13] Senge R , Coz J J D , E. Hüllermeier. Rectifying Classifier Chains for Multi-Label Classification. Space. Preprint arXiv:1906.02915, 2019.

[14] Cerri R , Mantovani R G , Basgalupp M P . Multi-label feature selection techniques for hierarchical multi-label protein function prediction. International Joint Conference on Neural Network (IJCNN'18), 2018.

[15] Wehrmann J , Barros R C . Hierarchical multi-label classification networks. International Conference on Machine Learning, pp.5075-5084, 2018.

[16] Huang S , Zhou Z . Multi-label learning by exploiting label correlations locally. Proceedings of the Twenty-Sixth AAAI Conference on Artificial Intelligence (AAAI'12), pp. 949-955, 2012.

[17] Johannes Fürnkranz, Eyke Hüllermeier, Eneldo Loza Mencía, et al. Multilabel classification via calibrated label ranking. Machine Learning, 73(2):133-153, 2008.

[18] Guo Y and Gu S. Multi-label classification using conditional dependency networks. In Proceedings of International Joint Conference on Artificial Intelligence (AAAI'11), pp.1300-1305, 2011.

[19] Huang J , Li G , Huang Q , et al. Learning Label-Specific Features and Class-Dependent Labels for Multi-Label Classification. IEEE Transactions on Knowledge & Data Engineering, 28(12):3309-3323, 2016.

[20] MinLing Zhang, Kun Zhang. Multi-label learning by exploiting label dependency. Proceedings of the 16th ACM SIGKDD International Conference on Knowledge Discovery and Data Mining (KDD'10), pp.999–1008, 2010.

[21] Ji M , Zhang K , Wu Q , et al. Multi-label learning for crop leaf diseases recognition and severity estimation based on convolutional neural networks. Soft Computing, 24:15327–15340, 2020.

[22] Read J , Pfahringer B , Holmes G , et al. Classifier chains for multi-label classification. Machine learning, 85(3): 333, 2011.

[23] Tsoumakas G , Katakis I , Vlahavas I . Random k-Labelsets for Multilabel Classification. IEEE Transactions on Knowledge & Data Engineering, 23(7):1079-1089, 2011.

[24] Loza Mencía, Eneldo, Janssen F . Learning rules for multi-label classification: a stacking and a separate-and-conquer approach. Machine Learning, 105(1):1-50, 2016.

[25] Boutell M R , Luo J , Shen X . Learning multi-label scene classification. Pattern Recognition, 37(9): 1757-1771, 2004.

[26] Zhang M L , Zhou Z H . ML-KNN: A lazy learning approach to multi-label learning. Pattern Recognition, 40(7):2038-2048, 2007.

[27] You M , Liu J , Li G. Embedded feature selection for multi-label classification of music emotions. International Journal of Computational Intelligence Systems, 5(4): 668-678, 2012.

[28] Wang S , Yuan L , Wang B , et al. Chinese Learner Feature Classification Based on IFE Attribute Weighted kNN Algorithm. Proceedings of 2020 IEEE 2nd International Conference on Computer Science and Educational Informatization (CSEI). IEEE, 2020.

[29] Wu F , Wang Z , Zhang Z , et al. Weakly Semi-Supervised Deep Learning for Multi-Label Image Annotation. IEEE Transactions on Big Data, 1(3):109-122, 2017.

[30] Wang Y , Zheng W , Cheng Y , et al. Joint label completion and label-specific features for multi-label learning algorithm. Soft Computing, 24:6553–6569, 2020.

[31] Zhang M L , Yukun L I , Liu X Y , et al. Binary relevance for multi-label learning: an overview. Frontiers of Computer ence, 12(2) , 2018.

[32] Cheng W , Hullermeier E , Dembczynski K J. Bayes optimal multilabel classification via probabilistic classifier chains. Proceedings of the 27th International Conference on Machine Learning (ICML'10), pp. 279-286, 2010.

[33] Zaragoza J H , Sucar L E , Morales E F , et al. Bayesian Chain Classifiers for Multidimensional Classification. Proceedings of International Joint Conference on Artificial Intelligence (IJCAI'11), AAAI Press, pp. 2192–2197, 2011.

[34] Dong H , Yang F and Wang X. Multi-label charge predictions leveraging label co-occurrence in imbalanced data scenario. Soft Computing, 2020.

[35] Pennington J , Socher R , Manning C. Glove: Global Vectors for Word Representation. Proceedings of Conference on Empirical Methods in Natural Language Processing (EMNLP'14), pp.1532-1543, 2014.

[36] Ponte J M , and Croft W B . A language modeling approach to information retrieval. In Proceedings of SIGIR'98, pp.275–281,1998.

[37] Madjarov G , Gjorgjevikj D , Delev T. Efficient two stage voting architecture for pairwise multi-label classification. Proceedings of Australasian Joint Conference on Artificial Intelligence (AI'10) ,pp.164-173, 2010.

[38] Huellermeier E , Fuernkranz J , Cheng W , et al. Label ranking by learning pairwise preferences. Artificial Intelligence, 172(16-17):1897-1916, 2008.